\documentclass{article}

    \usepackage[preprint]{neurips_2023}

\usepackage{amsmath,amsfonts,bm}

\def\eqref#1{equation~\ref{#1}}
\def\1{\bm{1}}

\DeclareMathAlphabet{\mathsfit}{\encodingdefault}{\sfdefault}{m}{sl}
\SetMathAlphabet{\mathsfit}{bold}{\encodingdefault}{\sfdefault}{bx}{n}
\newcommand{\tens}[1]{\bm{\mathsfit{#1}}}

\def\tF{{\tens{F}}}

\def\sS{{\mathbb{S}}}
\def\sT{{\mathbb{T}}}

\def\sX{{\mathbb{X}}}

\usepackage[utf8]{inputenc} % allow utf-8 input
\usepackage[T1]{fontenc}    % use 8-bit T1 fonts
\usepackage{hyperref}       % hyperlinks
\usepackage{url}            % simple URL typesetting
\usepackage{booktabs}       % professional-quality tables
\usepackage{amsfonts}       % blackboard math symbols
\usepackage{nicefrac}       % compact symbols for 1/2, etc.
\usepackage{microtype}      % microtypography
\usepackage{xcolor}         % colors
\usepackage{amsmath} 
\usepackage{adjustbox}
\usepackage{caption}
\usepackage{subcaption}
\usepackage{multirow}
\usepackage{graphicx}
\usepackage{float}
\usepackage{wrapfig, lipsum}
\usepackage{tabularx}
\usepackage{hhline}
\DeclareCaptionFont{myfont}{\fontsize{8pt}{11pt}\selectfont}

\title{Is Generative Modeling-based Stylization Necessary for Domain Adaptation in Regression Tasks?}

\author{%
  Jinman Park \\
  University of Waterloo\\
  Waterloo, ON, Canada \\
  \texttt{jinman.park@uwaterloo.ca} \\
  \And
  Francois Barnard \\
  University of Waterloo \\
  Waterloo, ON, Canada \\
  \texttt{fbarnard@uwaterloo.ca} \\
  \And
  Saad Hossain \\
  University of Waterloo \\
  Waterloo, ON, Canada \\
  \texttt{s42hossa@uwaterloo.ca} 
  \And
  Sirisha Rambhatla \\
  University of Waterloo \\
  Waterloo, ON, Canada \\
  \texttt{sirisha.rambhatla@uwaterloo.ca} 
  \And
  Paul Fieguth \\
  University of Waterloo \\
  Waterloo, ON, Canada \\
  \texttt{pfieguth@uwaterloo.ca} 
}
\usepackage{natbib}
\usepackage{subfiles}
\usepackage{xr}

\makeatletter
\newcommand*{\addFileDependency}[1]{% argument=file name and extension
  \typeout{(#1)}
  \@addtofilelist{#1}
  \IfFileExists{#1}{}{\typeout{No file #1.}}
}
\makeatother

\newcommand*{\myexternaldocument}[1]{%
    \externaldocument{#1}%
    \addFileDependency{#1.tex}%
    \addFileDependency{#1.aux}%
}

\myexternaldocument{supp}

\begin{document}

\maketitle

\begin{abstract}
  Unsupervised domain adaptation (UDA) aims to bridge the gap between source and target domains in the absence of target domain labels using two main techniques: input-level alignment (such as generative modeling and stylization) and feature-level alignment (which matches the distribution of the feature maps, e.g. gradient reversal layers). Motivated from the success of generative modeling for image classification, stylization-based methods were recently proposed for regression tasks, such as pose estimation. However, use of input-level alignment via generative modeling and stylization incur additional overhead and computational complexity which limit their use in real-world DA tasks. To investigate the role of input-level alignment for DA, we ask the following question: \emph{Is generative modeling-based stylization necessary for visual domain adaptation in regression?} Surprisingly, we find that input-alignment has little effect on regression tasks as compared to classification. Based on these insights, we develop a non-parametric feature-level domain alignment method -- Implicit Stylization (ImSty) -- which results in consistent improvements over SOTA regression task, without the need for computationally intensive stylization and generative modeling. Our work conducts a critical evaluation of the role of generative modeling and stylization, at a time when these are also gaining popularity for domain generalization.
\end{abstract}

\section{Introduction}
With the burst of interest and  applicability of deep learning applications in real-world settings \cite{liu2017survey, dong2021survey, pouyanfar2018survey}, there is a growing need to transfer knowledge from a source domain with labeled data to a target domain without labeled data. Unsupervised domain adaptation (UDA) has emerged \cite{s_uncertainty_2021,yang_multiple_2022, del_bimbo_adversarial_2021,  yan_augmented_nodate,chu_denoised_2022,dubourvieux_unsupervised_2021,huang_model_nodate} as a critical line of inquiry in machine learning since reliable labeling of datasets is often expensive and prohibitive. 

UDA can largely be divided into two levels of alignment: feature-level alignment \cite{kalischek2021light, zhou2021domain,chen2020homm,rozantsev2018beyond, kang2019contrastive, liu2020importance}, and input-level alignment \cite{stahlbock_brief_2021, yang_one-shot_2021,  zhou2022generative, wu2018dcan}. Feature-level alignment such as statistical moment matching \cite{long2017deep} and normalization statistics \cite{csurka2017domain, zhao2018unsupervised, zhao2020multi} aim to generate intermediate feature map representations that are similar in statistical distribution between the source and target domains, while input-level alignment such as domain style transfer \cite{sankaranarayanan2018generate, huang2017arbitrary} and adversarial learning \cite{ganin2016domain, liu2018data, he2020classification} aim to either stylize the source domain images in the style of target domain images or generate training data for the target domain. Generative modeling has shown promising results in image classification \cite{russo2018source} and to this day still shows SOTA results in the challenging MNIST $\to$ SVHN task. Success of input-alignment has translated over to regression task such as pose estimation \cite{kim2022unified} where stylization \cite{huang2017arbitrary} was employed. However, given the computational overhead of generative methods with input-level alignment, it is still unclear in literature if generative methods are more effective than feature-level alignment methods.  

\begin{figure}[t!]
    \centering
    % \begin{subfigure}{0.45\textwidth}
    %     \centering
    %     \includegraphics[width=\textwidth]{Figures/pose_comparison_uda-macs.drawio.pdf}
    %         \caption[MACs]%
    %         {{\small MACs}} 
    %     \end{subfigure}
    %     \begin{subfigure}{0.45\textwidth}  
    %         \centering 
    %         \includegraphics[width=\textwidth]{Figures/pose_comparison_uda-params.drawio.pdf}
    %         \caption[Parameters]%
    %         {{\small Parameters}}   
    %     \end{subfigure}

    \includegraphics[width=0.7\textwidth]{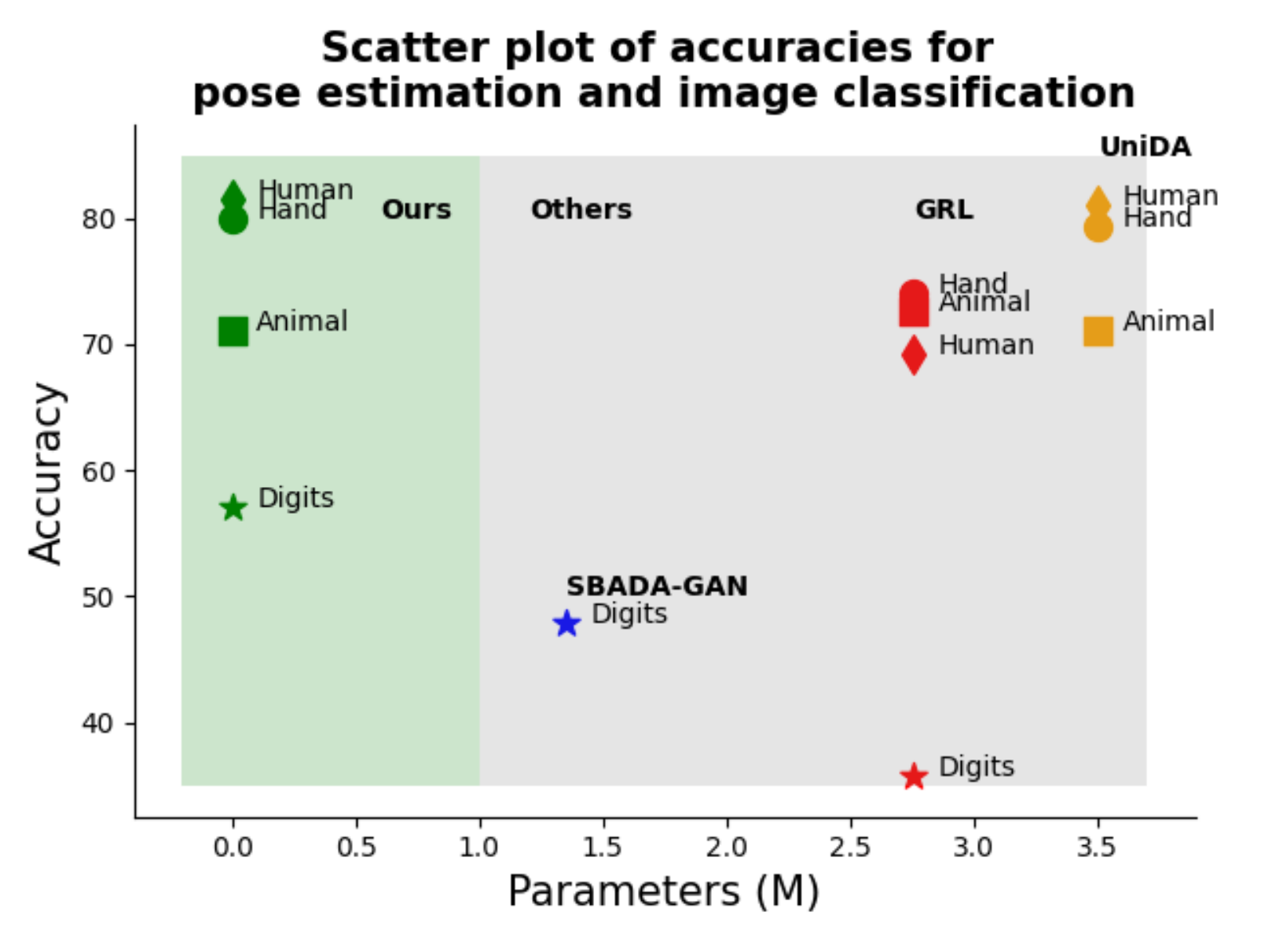}
    \caption{ \textbf{Comparison of pose estimation and image classification accuracy for target domain respective to the number of trainable parameters used in domain alignment.} Our method achieves SOTA results in both image classification and pose estimation while significantly reducing the number of trainable parameters for domain alignment used in UDA. \emph{Hand, Human, \& Animal} refer to pose estimation while \emph{Digits} refers to image classification \ref{app:otherexp}.}
    \label{fig:scatter}
     \vspace{-10pt}
\end{figure}

Given this gap in research, our key observation is that input-level alignment via generative models and stylization are computationally expensive and in some cases difficult to train in data scarce regimes. Consequently, a natural question emerges based on this observation: \emph{Is generative modeling-based stylization necessary for visual domain adaptation in regression}?  

To begin addressing this question, we investigate the role of input-alignment and show that it does not consistently improve the results as opposed to a no-stylization baseline with a mean-teacher training scheme. Moreover, we propose a computationally efficient implicit stylization method, \textbf{ImSty}. Building on AdaIN, the concept of adaptive instance normalization \cite{huang2017arbitrary} blocks, we incorporate AdaIN blocks into downstream tasks of pose estimation where the mini-batch level statistics of source domain and target domain are swapped. Our method avoids the need for explicitly training a generative model and generating input data in the style of the target domain, saving significant amounts of inefficient computations and training time. 

We evaluate our proposed implicit stylization method on three pose estimation datasets with different levels of domain gaps: Rendered Hand Pose Dataset \cite{zimmermann2017learning} $\rightarrow$ Hand-3D-Studio \cite{zhao2020hand}, SURREAL \cite{varol2017learning} $\rightarrow$ Leeds Sports Pose \cite{johnson2010clustered}, and Synthetic Animal Dataset \cite{mu2020learning} $\rightarrow$ TigDog Dataset \cite{del2015articulated}. We achieve SOTA results on all three datasets while reducing the number of trainable parameters by 100\% and the number of computations (MACs) by 99.99\% compared against the explicit stylization used in \citet{kim2022unified}. %In addition, we experiment on MNIST \cite{deng2012mnist} $\rightarrow$ SVHN \cite{netzer2011reading}, which is known for being notoriously difficult \cite{french2017self, shu2018dirt, dai2020contrastively, kumar2018co} without specific data augmentations such as intensity flipping and standardization. We show that with implicit stylization, we achieve SOTA results with minimal data augmentation and no specific data augmentation to reflect real-world scenarios where it can be costly and time consuming to find the right data augmentation that works for the target domain. Code to reproduce the results is provided in the supplementary material.

\section{Related Works}

\subsection{Unsupervised Domain Adaptation}
Unsupervised Domain Adaptation (UDA) \cite{borgwardt2006integrating, huang2006correcting} provides a suitable solution for data scarce domains. The aim of UDA is to associate a given labeled source domain, with abundant data, to an unlabeled target domain. Prevailing approaches for UDA often utilize adversarial learning methods \cite{s_uncertainty_2021,yang_multiple_2022, del_bimbo_adversarial_2021},  pseudo-labeling methods \cite{yan_augmented_nodate,chu_denoised_2022,dubourvieux_unsupervised_2021,huang_model_nodate}, or both as a foundation. These methods allow for a wide variety of focused implementations that provide reliable adaptation results. As proposed by \citet{del_bimbo_adversarial_2021}, the application of adversarial methods produces more resistant and robust models. They describe an adversarial continuous learning in unsupervised domain adaptation (ACDA) method that yields a relative error reduction of 20\% \cite{del_bimbo_adversarial_2021}. Similarly, pseudo-labeling methods \cite{choi_pseudo-labeling_2019, lee2013pseudo} are commonplace in novel UDA works as they provide solutions for data scarcity.

\subsection{Input-level Alignment (Generative methods)}
Various attempts aim to remedy scarce source datasets with variations of domain adaptation (DA). Of these numerous options, there have been notable developments in using adversarial or generative models alongside DA techniques. As described by \citet{stahlbock_brief_2021}, the use of generative adversarial networks (GANs) was inspired by their ability to minimize the domain shift commonplace in DA. The fundamental notion of these techniques is to generate representative data of the source domain that can be used for further training and fine-tuning. With the development of their model, GenDA, \citet{yang_one-shot_2021} propose an alternative approach to generative domain adaptation (GDA) applications. By freezing the parameters of a pre-trained GAN, \citet{yang_one-shot_2021} reuse the prior information from the source GAN model to adapt and generate new content in the target domain. Moreover, novel GDA approaches \citet{zhou2022generative} and \citet{wu2018dcan} improve generalizability and reduce gaps within the domain by leveraging neural consistency in statistics to guide the generator, incorporating dual-level semantic consistency, while proposing intra-domain spectrum mixup.

\subsection{Feature-level Alignment}
In addition to alignment at the input level, methods involving alignment at the feature level have been explored in numerous ways. One such pathway involves methods centered around the use of divergence measures such as MixStyle, MMD, CORAL, CDD and Wasserstein Distance, which all share the goal of minimizing domain discrepancy at the feature level \cite{kalischek2021light, zhou2021domain,chen2020homm,rozantsev2018beyond, kang2019contrastive, liu2020importance}. Furthermore, studies have also employed adversarial methods \cite{bousmalis2017unsupervised} generally comprised of three modules: a feature classifier, discriminator, and label classifier. In contrast, methods such as \citet{liu2021adversarial} iteratively add a label distribution estimator module to attempt alignment accounting for both label and domain. Moreover, DANN \cite{ganin2016domain} opts to use a gradient reversal layer to allow for domain discrimination. There have also been attempts to utilize batch-normalization (BN) statistics in UDA \cite{li2018adaptive, maria2017autodial,chang2019domain} showing that BN statistics encapsulate domain data.

Over time, both input-level alignment and feature-level alignment have shown tremendous amount of success in DA. However, given the high computational cost of input-level alignment, there exists a gap in research that explicitly compares input-level alignment and feature-level alignment. To fill this gap in research, and to question if input-level alignment is necessary in domain adaptation, we propose a feature-level alignment method called ImSty that will be discussed in the following section.

\section{Implicit Stylization}
In this section, we develop a feature-level alignment method called implicit stylization (\textbf{ImSty}) that replaces generative modeling (i.e., explicit stylization or generative models) in various domain adaptation tasks. The main idea is to incorporate adaptive instance normalization into the training pipeline to merge domain gaps without having the need of a generative model with a full auto-encoder structure for pixel-to-pixel generation. Our method does not require training an explicit stylization model or require any additional trainable parameters. 

\subsection{Implicit Stylization in Pose Estimation}
\citet{kim2022unified} achieved SOTA results in unsupervised domain adaptation for 2D pose estimation by unifying a framework that generalizes well on various poses with different levels of input and output variance. To merge the gap between the source and target domains, the authors proposed stylizing the source images to target images for the student model, and stylizing the target images to the source images for the teacher model. However, the explicit stylization requires one to train a stylization model and generate new images every batch, leading to computational inefficiency in training. Therefore, to mitigate inefficiency, we propose extracting the adaptive instance normalization block from the explicit stylization model and directly applying it to the encoders for pose estimation.

Given a labeled pose dataset from the source domain $\sS$ that contains images $\sX_{s} \in \mathbb{R}^{H \times W \times 3}$ and corresponding keypoint labels $y_{s} \in \mathbb{R}^{K \times 2}$, along with an unlabeled pose dataset from the target domain $\sT$ with just images $\sX_{t}\in \mathbb{R}^{H \times W \times 3}$, the goal is to generalize a model $h$ to the target domain $\sT$ based on learning from the source domain $\sS$. %Generally, instead of training the model to predict keypoint coordinates, heatmaps $H = L(y) \in \mathbb{R}^{K \times H' \times W'}$ generated by $L : \mathbb{R}^{K \times 2} \rightarrow \mathbb{R}^{K \times H' \times W'}$ are used as labels to train the model by minimizing the MSE loss function 

\begin{figure*}[t]
    \centering
    \begin{tabular}{cc}
     \multicolumn{2}{c}{\includegraphics[width=\textwidth]{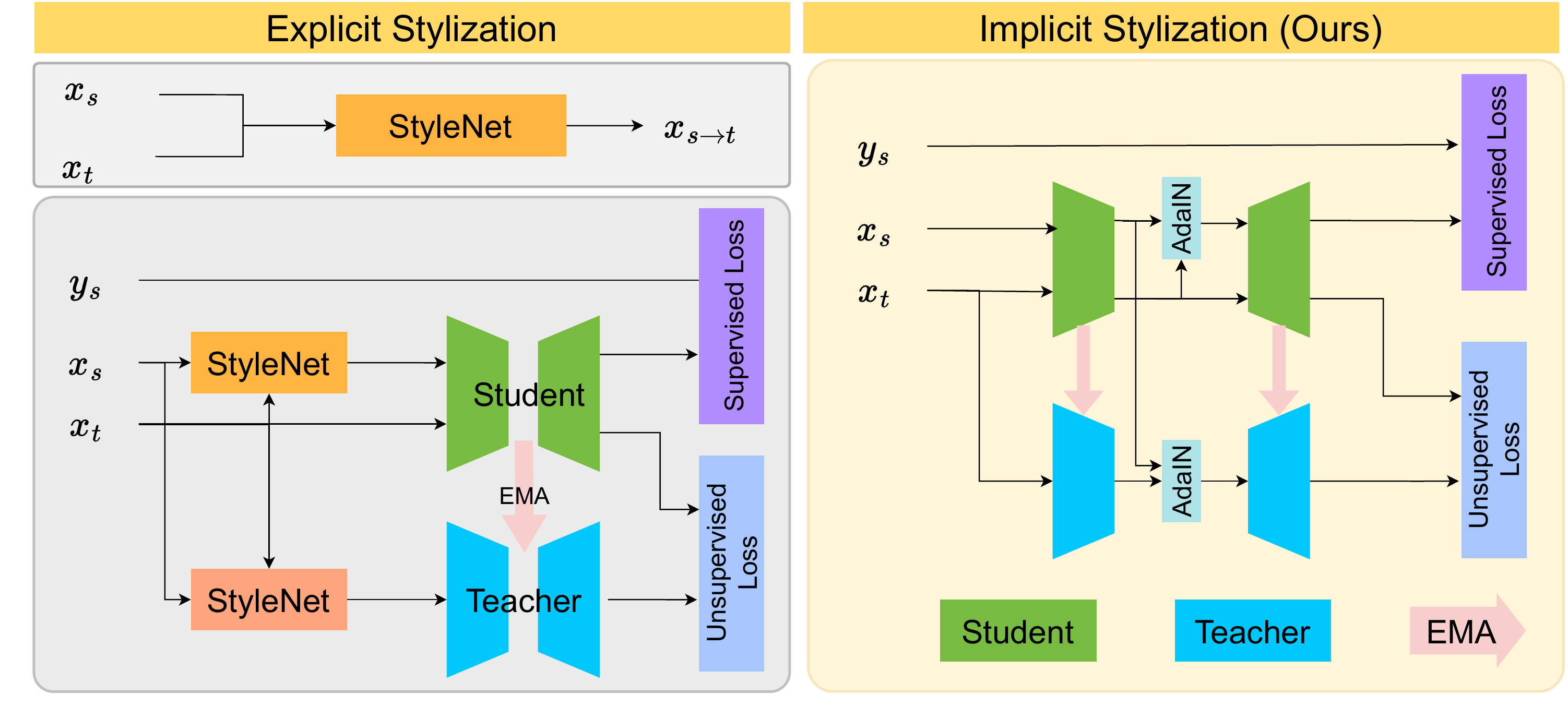}}      \\
      { \hspace{30pt} (a) Explicit stylization of \citet{kim2022unified}}   & { (b) Proposed (Implicit stylization)}\\
    \end{tabular}
    % \begin{subfigure}{0.45\textwidth}
    %     \centering
    %     \includegraphics[width=\textwidth]{Figures/uda_comparison-Page-1.drawio.pdf}
    %         \caption[Explicit stylization]%
    %         {{\small Explicit stylization in Kim et al. \cite{kim2022unified}}} 
    %     \end{subfigure}
    %     \hspace{5pt}
    %     % \hfill
    %     \begin{subfigure}{0.45\textwidth}  
    %         \centering 
    %         \includegraphics[width=\textwidth]{Figures/uda_comparison-Page-2.drawio.pdf}
    %         \caption[Ours (Implicit stylization)]%
    %         {{\small Ours (Implicit stylization)}}   
    %     \end{subfigure}     
        \caption[]{\textbf{Comparison between explicit stylization and our proposed implicit stylization.} Explicit stylization (left) pre-trains a stylization model (e.g., StyleNet or neural style transfer model) before a downstream task. Then, the style transfer model is used to generate batches of source domain images stylized in the target domain, and target domain images stylized in source domain. In comparison, our proposed implicit stylization (right) method does not require any extra models (trainable parameters) and only requires a small amount of computation for the statistics in \eqref{eq:normalization}.}
    \label{fig:modeldiagram}
\end{figure*}

Now, we describe the details of our proposed implicit stylization method. A mini-batch size of $n \ll N_{s} \text{ and } n \ll N_{t}$ are sampled from both $\sS$ and $\sT$. Given a set of source domain images $\sX_{s} = \{x^{1}_{s}, x^{2}_{s},\dots,x^{n}_{s} \}$ and a set of target domain images $\sX_{t} = \{x^{1}_{t},x^{2}_{t},\dots,x^{n}_{t} \}$, $\sX_{s}$ and $\sX_{t}$ are passed through a ResNet-101 \cite{he2016deep} backbone $R$ to obtain feature maps $\tF_{s} = R(\sX_{s}), \tF_{t} = R(\sX_{t})$. Each channel of $C$ in $\tF_{s} \in \mathbb{R}^{H' \times W' \times C}$ and $\tF_{t} \in \mathbb{R}^{H' \times W' \times C}$ is normalized to $\mathcal{N}(0,1)$:% while keeping the mean $\mu \in \mathbb{R}^{C}$ and standard deviation $\sigma \in \mathbb{R}^{C}$ of each channel used for the normalization for $j \in \{ s, t\}$:
\begin{equation}
\begin{aligned}
    \mu^{i}_{j} = \texttt{Mean}(\tF_{j}[:, :, i]), ~~
    \sigma^{i}_{j} = \texttt{STD}(\tF_{j}[:, :, i])&, ~~
     \mu_{j} = [\mu^{1}_{j}, \mu^{2}_{j}, \dots, \mu^{C}_{j}], ~~
     \sigma_{j} = [\sigma^{1}_{j}, \sigma^{2}_{j}, \dots, \sigma^{C}_{j}] \\
     \tF_{\text{norm}, s} = \frac{\tF_{s} - \mu_{s}}{\sigma_{s}}&, ~~
     \tF_{\text{norm},t} = \frac{\tF_{t} - \mu_{t}}{\sigma_{t}}
    \end{aligned}
    \label{eq:normalization}
\end{equation}
Then, the mean and standard deviation from the opposite domain are utilized to reverse the normalization, building on the concept of AdaIN \cite{huang2017arbitrary}: 
\begin{equation}
\begin{aligned}
    \tF_{s \rightarrow t} &= \alpha(f_{\text{norm},s}\cdot \sigma_{t} +\mu_{t}) + (1-\alpha)\tF_{s} \\
     \tF_{t \rightarrow s} &= \alpha(f_{\text{norm},t}\cdot \sigma_{s} +\mu_{s}) + (1-\alpha)\tF_{t}
    \end{aligned}
    \label{eq:reversenormalization}
\end{equation}
$\tF_{s \rightarrow t}$ and $\tF_{t}$ are passed to the student decoder $D_{\text{stu}}$
\begin{equation}
\begin{aligned}
    \hat{y}_{s} = D_{\text{stu}}(\tF_{s \rightarrow t}), \hspace{10pt}
     \hat{y}_{t} = D_{\text{stu}}(\tF_{t}) 
    \end{aligned}
    \label{eq:decoder}
\end{equation}
\begin{table*}[t]
    \centering
    \begin{center}
    \caption{\textbf{Pose Estimation Comparison for hand pose, animal pose, and human pose estimation.} Our proposed ImSty method achieves SOTA results in pose estimation across three different domains (hand pose, human pose, and animal pose). In hand pose and human pose we achieve the highest accuracy while achieving competitive results in animal pose. In the table, the superscripts ``${P}$'' and ``${R}$'' denote the published metrics (reported in the papers) and the reproduced values, respectively. For hand pose estimation, MCP, PIP, DIP are acryonyms for metacarpophalangeal, proximal
interphalangeal, and distal interphalangeal respectively.}
    \label{tab:mainresults}
    \end{center}
    \vspace{-5pt}
    \newcommand{\w}{40pt}
    \resizebox{\textwidth}{!}{\begin{tabular}{|m{120pt}|m{\w}m{\w}m{\w}m{\w}m{\w}|}
        \hline
        \multicolumn{6}{|c|}{\textbf{Rendered Hand Pose Dataset $\to$Hand-3D-Studio }} \\
        \hhline{|======|}
        \centering Method                  & MCP           & PIP           & DIP           & Fingertip     & All                \\ 
        \hline
        \centering CCSSL \cite{mu2020learning} & 81.5          & 79.9          & 74.4          & 64.0          & 75.1           \\
        \centering \footnotesize{UDA-Animal \cite{li2021synthetic}} & 82.3          & 79.6          & 72.3          & 61.5          & 74.1     \\
        \centering RegDA \cite{jiang2021regressive} & 79.6          & 74.4          & 71.2          & 62.9          & 72.5      \\
        \centering UniDA$^{P}$ \cite{kim2022unified} & 86.7          & 84.6          & 78.9          & 68.1          & 79.6     \\
        \centering UniDA$^{R}$\cite{kim2022unified}            & 87.1          & 85.1          & \textbf{79.3} & 68.5          & 79.9               \\
        \hline
        \centering Ours (ImSty)   & \textbf{87.6} & \textbf{85.5} & 78.9          & \textbf{68.7} & \textbf{80.1}                     \\ 
        \hline
    \end{tabular}}
    \newcommand{\y}{20pt}
    \resizebox{\textwidth}{!}{\begin{tabular}{|m{120pt}|m{\y}m{\y}m{\y}m{\y}m{\y}m{\y}m{\y}m{\y}|}
        \multicolumn{1}{|c}{}   &               &               &               &               &               &               &               &                \\ 
        \hline
        \multicolumn{9}{|c|}{\textbf{Synthetic Animal Dataset $\to$ TigDog Dataset}}                                                                             \\ 
        \hhline{|=========|}
        \centering \multirow{2}{*}{Method} & \multicolumn{8}{c|}{Horse}                                                                                                     \\
        \cline{2-9}
         \centering              & Eye           & Chin          & Hoof          & Hip           & Knee          & Shd      & Elbow         & All            \\ 
        \hline
        \centering CCSSL \cite{mu2020learning} & 89.3          & 92.6          & 65.0          & 78.1          & 73.1          & 69.5          & 70.0          & 73.1           \\
        \centering \footnotesize{UDA-Animal \cite{li2021synthetic}} & 86.9          & \textbf{93.7} & \textbf{72.6} & 81.9          & \textbf{79.1} & \textbf{76.4} & 70.6          & \textbf{77.5}  \\
        \centering RegDA \cite{jiang2021regressive} & 89.2          & 92.3          & 63.2          & 77.5          & 72.7          & 70.5          & 71.5          & 73.2           \\
        \centering UniDA$^{P}$ \cite{kim2022unified} & 91.3          & 92.5          & 66.6          & 74.2          & 77.0          & 74.0          & \textbf{75.8} & 76.4           \\
        \centering UniDA$^{R}$ \cite{kim2022unified}            & 91.5          & 93.6          & 67.2          & \textbf{82.3} & 77.0          & 73.1          & 74.8          & 75.6           \\
        \hline
        \centering Ours (ImSty)                   & \textbf{91.6} & 92.8          & 66.2          & 77.1          & 76.5          & 72.6          & 74.7          & 75.4           \\ 
        \hline
        \multicolumn{1}{|c}{}   &               &               &               &               &               &               &               &                \\ 
        \hline
        \centering \multirow{2}{*}{Method} & \multicolumn{8}{c|}{Tiger}                                                                                                     \\ 
        \cline{2-9}
        \centering & Eye           & Chin          & Hoof          & Hip           & Knee          & Shd      & Elbow         & All            \\ 
        \hline
        \centering CCSSL \cite{mu2020learning} & 94.3          & 91.3          & 70.2          & 70.2          & 59.1          & 49.5          & \textbf{53.9} & 66.7           \\
        \centering \footnotesize{UDA-Animal \cite{li2021synthetic}} & 98.4          & 87.2          & \textbf{73.4} & \textbf{74.9} & 62.0          & 49.4          & 49.8          & 67.7           \\
        \centering RegDA \cite{jiang2021regressive} & 93.3          & 92.8          & 60.7          & 67.8          & 55.4          & 50.3          & 50.2          & 61.8           \\
        \centering UniDA$^{P}$ \cite{kim2022unified} & \textbf{98.5} & 96.9          & 72.8          & 63.7          & \textbf{62.8} & \textbf{56.2} & 52.3          & \textbf{67.9}  \\
        \centering \footnotesize{UniDA$^{R}$ \cite{kim2022unified}}           & 98.3          & \textbf{97.0} & 72.1          & 71.7          & 62.2          & 53.4          & 53.0          & 67.1           \\ 
        \hline
        \centering Ours (ImSty)                   & 97.7          & 96.3          & 72.1          & 72.5          & 61.3          & 52.9          & 52.2          & 66.9           \\ 
        \hline
        \multicolumn{1}{|c}{}   &               &               &               &               &               &               &               &                \\ 
        \hline
    \end{tabular}}
    \newcommand{\human}{25pt}
    \resizebox{\textwidth}{!}{\begin{tabular}{|m{120pt}|m{\human}m{\human}m{\human}m{\human}m{\human}m{\human}m{\human}|}
        \multicolumn{8}{|c|}{\textbf{SURREAL $\to$ Leeds Sports Pose}}                                                                      \\ 
        \hhline{|========|}
        \centering Method                 & Shd      & Elbow         & Wrist         & Hip           & Knee          & Ankle         & All       \\ 
        \hline
        \centering CCSSL \cite{mu2020learning} & 36.8          & 66.3          & 63.9          & 59.6          & 67.3          & 70.4          & 60.7      \\
        \centering \footnotesize{UDA-Animal \cite{li2021synthetic}} & 61.4          & 77.7          & 75.5          & 65.8          & 76.7          & 78.3          & 69.2      \\
        \centering RegDA \cite{jiang2021regressive} & 62.7          & 76.7          & 71.1          & 81.0          & 80.3          & 75.3          & 74.6      \\
        \centering UniDA$^{P}$ \cite{kim2022unified} & 69.2          & 84.9          & 83.3          & 85.5          & 84.7          & 84.3          & 82.0      \\
        \centering UniDA$^{R}$ \cite{kim2022unified}            & 68.4          & \textbf{85.6} & 83.1          & \textbf{86.2} & 85.0          & 84.2          & 82.0      \\ 
        \hline
        \centering Ours (ImSty)                   & \textbf{71.1} & 85.2  & \textbf{83.4} & 85.2          & \textbf{86.1} & \textbf{85.1} & \textbf{82.6}     \\
        \hline
    \end{tabular}}
    \vspace{-10pt}
\end{table*}
that up-samples the feature maps to $\hat{y}_{s}\in \mathbb{R}^{h \times w \times J}$ and $\hat{y}_{t}\in \mathbb{R}^{h \times w \times J}$ with height $h$, width $w$, and number of joints $J$. $\alpha$ is sampled randomly from a Uniform(0,1) distribution. Then, $\tF_{t \rightarrow s}$ is passed to the teacher decoder $D_{\text{tea}}$ to generate pseudo-labels $y_{t} = D_{\text{tea}}(\tF_{t \rightarrow s})$. 

Mean-squared error (MSE) metric is used as the loss function for both supervised $\mathcal{L}_{\text{sup}}$ and unsupervised $\mathcal{L}_{\text{unsup}}$ loss with a weighting factor $\lambda$ for the overall loss $\mathcal{L}_{\text{total}}$:
\begin{equation}
\begin{aligned}
    \mathcal{L}_{\text{sup}} = \text{MSE}(y_{s}, \hat{y}_{s }), \hspace{10pt}
    \mathcal{L}_{\text{unsup}} = \text{MSE}(y_{t}, \hat{y}_{t}),
    \hspace{10pt}
    \mathcal{L}_{\text{total}} = \mathcal{L}_{\text{sup}}+\lambda \mathcal{L}_{\text{unsup}}
    \end{aligned}
    \label{eq:loss}
\end{equation}

Remaining details on the overall mean-teacher pipeline, such as pseudo-label masking, normalization, adaptive occlusions, and reverse augmentations, can be found in \citet{kim2022unified}. Refer to Figure~\ref{fig:modeldiagram} for a comparison between explicit stylization and our proposed implicit stylization method.

\section{Experiments}
In this section, we begin to incrementally reveal the excessive nature of generative methods for UDA by showing either on-par or better results on pose estimation datasets with close to zero additional computations and with zero additional trainable parameters. We now describe the experimental set-up to evaluate the proposed method as compared to the current state-of-the-art. We also release the code corresponding to our experiments for reproducibility. Additional experiments on digits classification are listed in Appendix~\ref{app:otherexp}.

\begin{table*}[]
    \centering
    \caption{\textbf{Comparison of trainable parameters and computational costs (MACs) for merging domain gaps.} Our implicit stylization method requires no additional trainable parameters and minimal computations to calculate feature-level statistics to merge the gap between source and target domains. For fair comparison, all methods use a fixed image resolution of $256 \times 256$. }
    \begin{tabular}{cccc}
    \toprule 
         Method & Params. (M) & MACs (G) & Generative\\
         \hline 
         %SBADA-GAN \cite{russo2018source} & 51.73 & 42.54 & \checkmark \\
         StyleNet \cite{huang2017arbitrary} & 3.51 & 63.11 & \checkmark\\
         RevGrad \cite{ganin2015unsupervised} & 2.75 & 0.34\\
         Ours (ImSty) &\textbf{ 0.00} & \textbf{1.84e-3} \\
         \bottomrule
    \end{tabular}
    \label{tab:computation}
\end{table*}
\begin{figure*}[h]
    \centering
    \begin{subfigure}{0.16\textwidth}
        \centering
        \includegraphics[width=\textwidth]{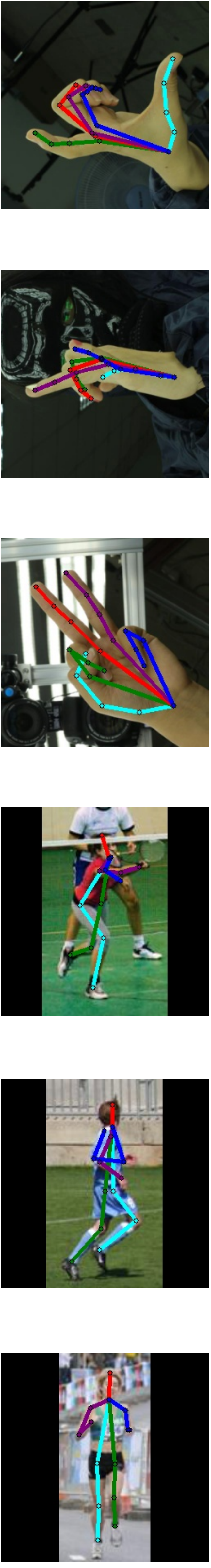}
            \caption[Ground Truth]%
            {{\small Ground Truth }} 
        \end{subfigure}
        \hspace{10pt}
        \begin{subfigure}{0.16\textwidth}  
            \centering 
            \includegraphics[width=\textwidth]{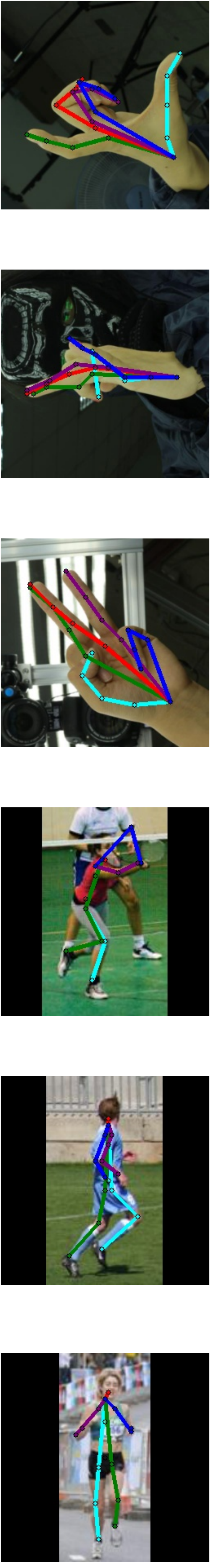}
            \caption[No Sty.]%
            {{\small No Sty.}}   
        \end{subfigure}
        \hspace{10pt}
        \begin{subfigure}{0.16\textwidth}  
            \centering 
            
            \captionsetup{font=myfont}
            \includegraphics[width=\textwidth]{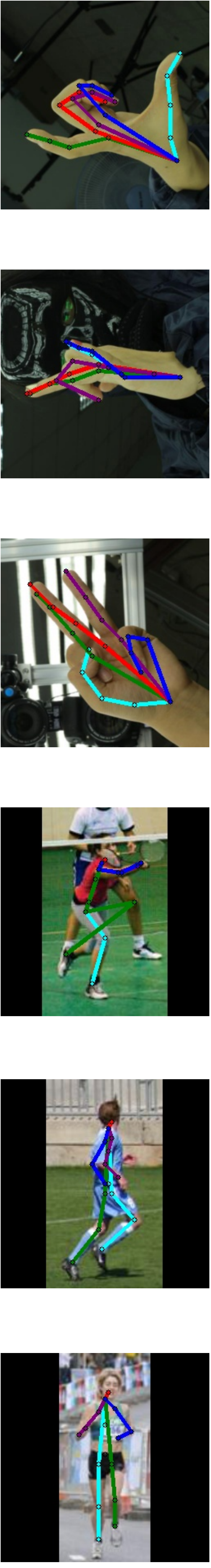}
            \vspace{-12pt}
            \caption[]%
            { \cite{kim2022unified}}   
        \end{subfigure} 
        \hspace{10pt}
        \begin{subfigure}{0.16\textwidth}  
            \centering 
            \includegraphics[width=\textwidth]{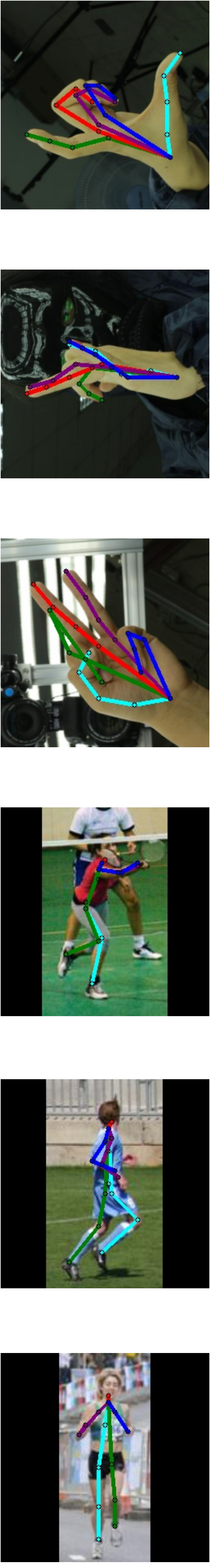}
            \caption[Ours]%
            {{\small Ours (ImSty)}}   
        \end{subfigure}  
        \caption[]{\textbf{Qualitative evaluation on pose estimation.} There are two main observations to be made in this figure. First, column (b) still shows competitive results despite not having any domain alignment. Second, column (d) shows the best qualitative results. These two observations suggest that explicit input-alignment via stylization might not be necessary in pose estimation. Note that column (b) is \cite{kim2022unified} without stylization.} 
    \label{fig:qualitativeeval}
    \vspace{-10pt}
\end{figure*}
% \begin{table*}[t]
%     \centering
%     \caption{\textbf{Digit classification results on MNIST $\rightarrow$ SVHN.} It is well established from \cite{french2017self} that MNIST $\rightarrow$ SVHN is much more challenging than SVHN $\rightarrow$ MNIST. In addition, the results of many works vary greatly depending on the specific data augmentation. Since manually searching for data augmentations that work for the target domain is costly and time-consuming, we compare SOTA methods on minimal data augmentations. Observe that given minimal data augmentations, our method achieves SOTA results on UDA digit classification without generative methods. "SBADA-GAN$^{R}$" indicates reproduced values.}
%     \begin{tabular}{ccc}
%     \toprule 
%          Method & Accuracy & Generative\\
%          \midrule 
%          RevGrad \cite{ganin2015unsupervised} & 35.7 & \\
%          DCRN \cite{ghifary2016deep} & 40.1 &\\
%          G2A \cite{sankaranarayanan2018generate} & 36.4 & \checkmark\\
%          SE \cite{french2017self} & 37.5 &\\
%          ATT \cite{saito2017asymmetric} & 52.8 &\\
%          SBADA-GAN$^{R}$ \cite{russo2018source} & 47.9 $\pm$ 1.7&\checkmark\\
%          % SBADA-GAN\cite{russo2018source} & \textbf{61.1} &\checkmark\\
%          PFAN \cite{chen2019progressive} & 57.6 $\pm$ 1.8 &\\
%          DIRT-T \cite{shu2018dirt} & 54.5 & \\
%          Ours (ImSty) & 	\textbf{57.82} $\pm$ 3.2 & \\
%          \bottomrule
%     \end{tabular}
%     \label{tab:digits}
% \end{table*}

\begin{figure*}
\centering
\begin{subfigure}{0.4\textwidth}
\includegraphics[width=\textwidth]{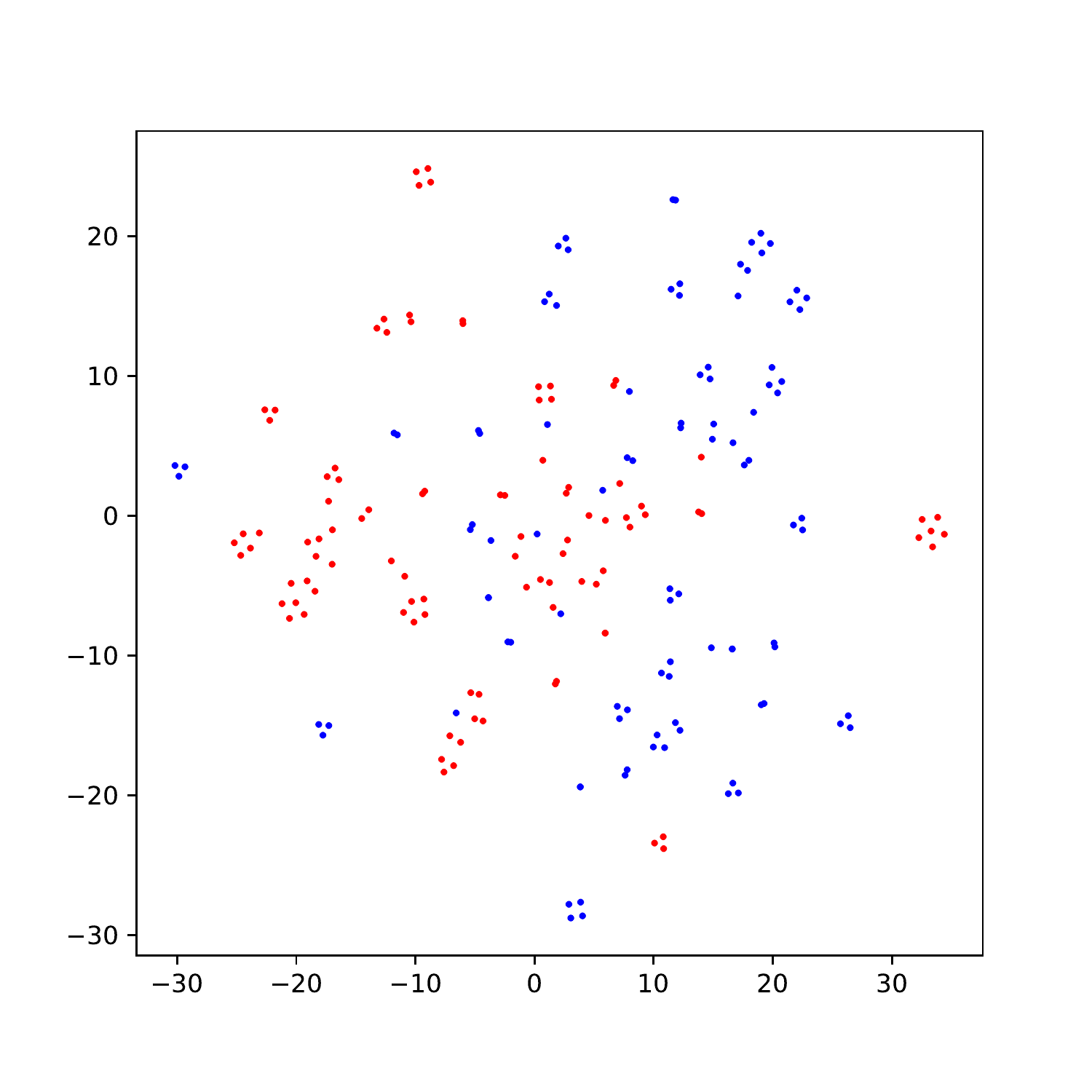}
\caption{Ours (Hand)}
\end{subfigure}
\hspace{0cm}
\begin{subfigure}{0.4\textwidth}
\includegraphics[width=\textwidth]{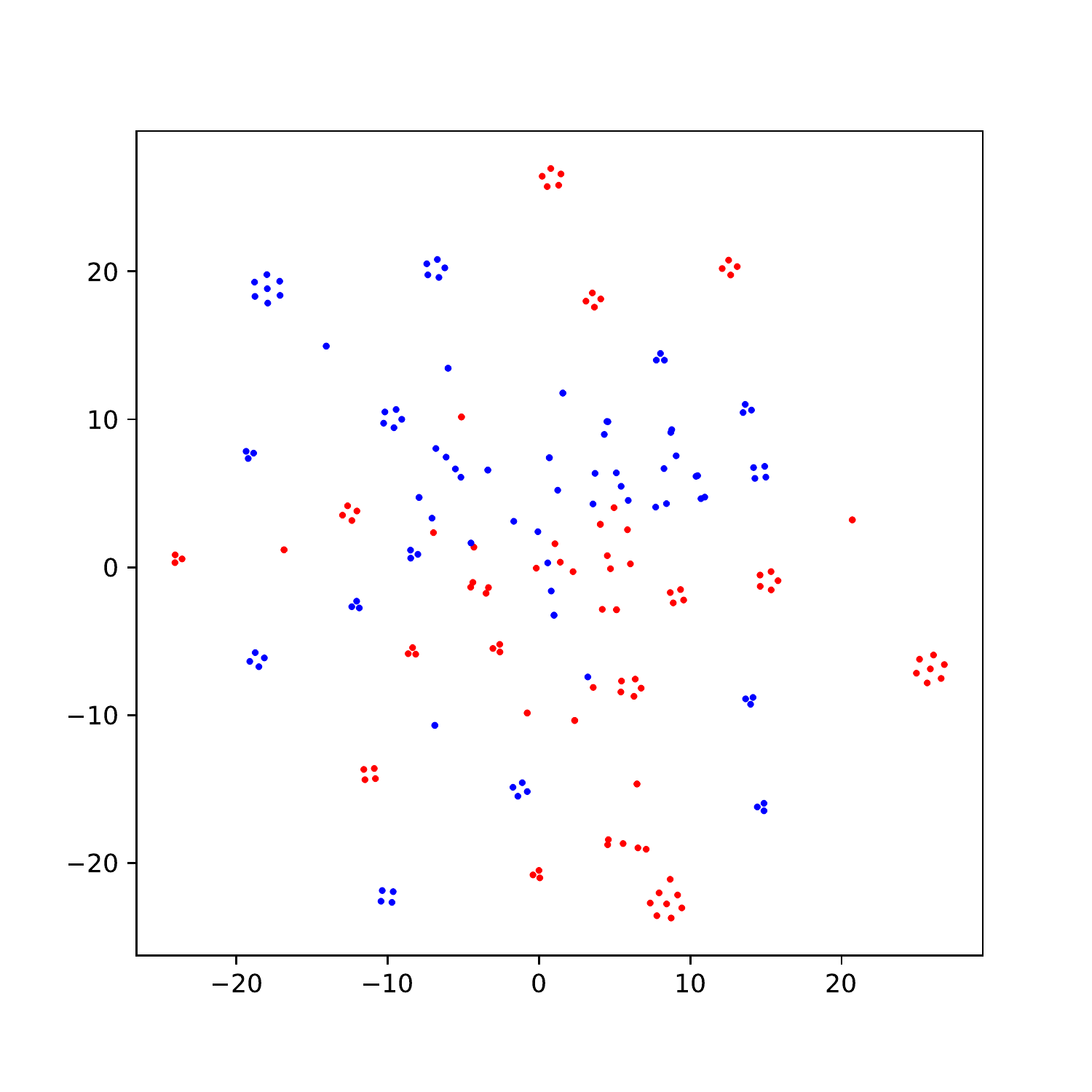}
\caption{Generative \cite{kim2022unified} (Hand)}
\end{subfigure}
\vspace{-20pt}\vskip\baselineskip
\begin{subfigure}{0.4\textwidth}
\includegraphics[width=\textwidth]{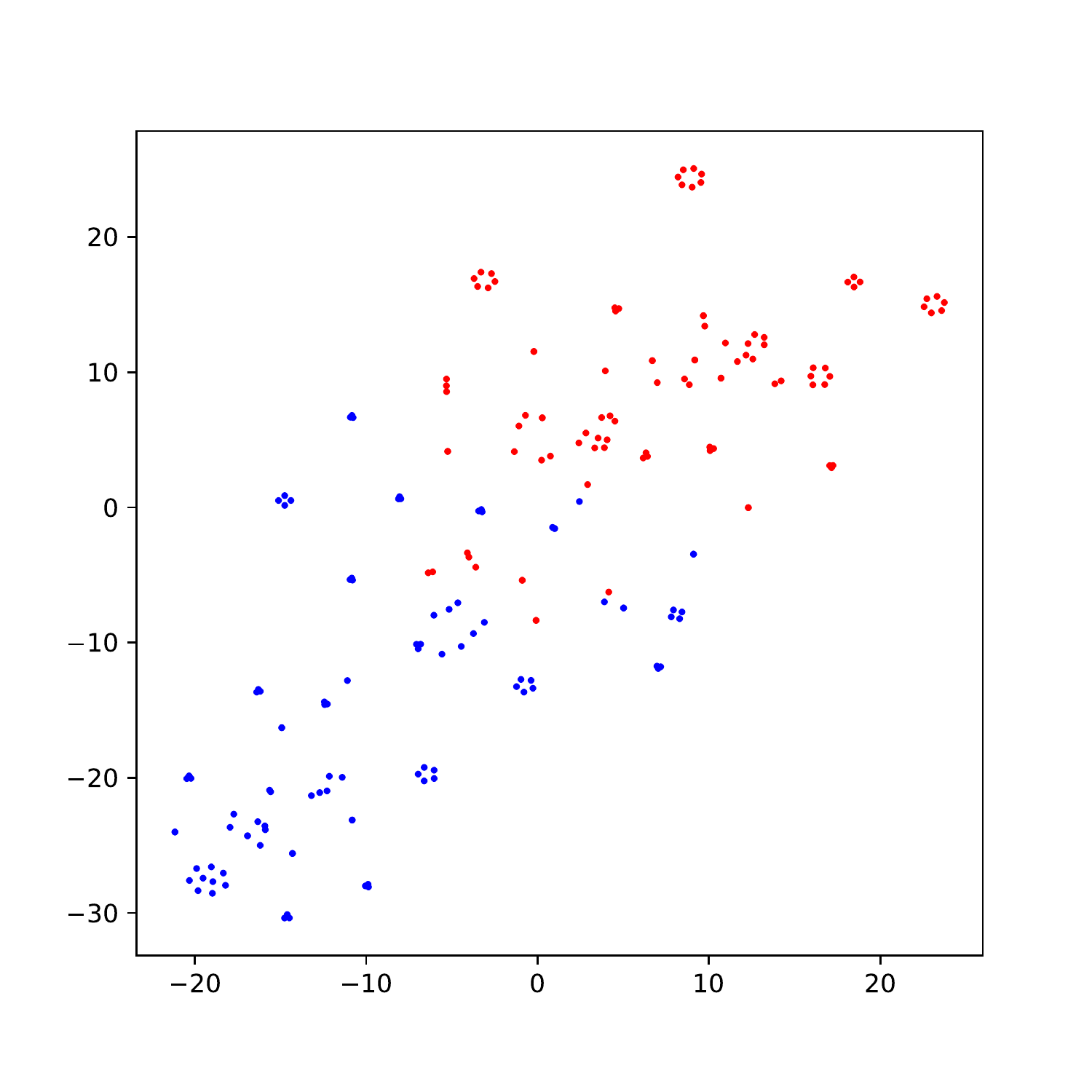}
\caption{Ours (Human)}
\end{subfigure}
\hspace{0cm}
\begin{subfigure}{0.4\textwidth}
\includegraphics[width=\textwidth]{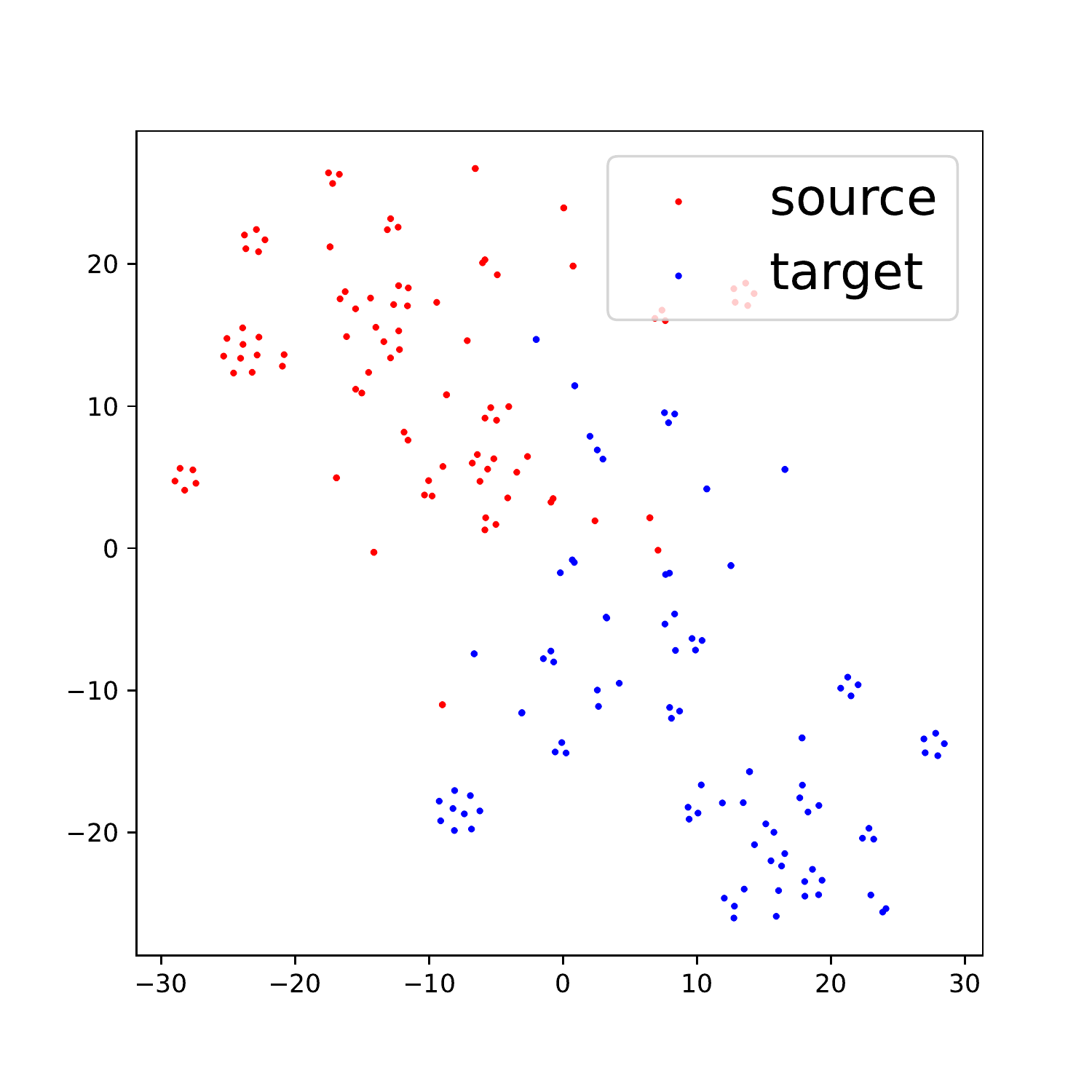}
\caption{Generative \cite{kim2022unified}(Human)}
\end{subfigure}

\caption{\textbf{TSNE comparison between    \cite{kim2022unified} and our proposed ImSty method.} Visual inspection of the TSNE clustering between the input stylization in \cite{kim2022unified} demonstrates close to no difference with our proposed ImSty method. This suggests that ImSty can achieve similar domain alignment compared to the input stylization used in \cite{kim2022unified}.}
\label{fig:tsnediagram}
\vspace{-13pt}
\end{figure*}

\subsection{Datasets}

We evaluate our implicit stylization pipeline using three sets of commonly used UDA pose estimation datasets with high variances in both input and output settings: \textbf{Rendered Hand Pose Dataset} \cite{zimmermann2017learning} $\rightarrow$ \textbf{Hand-3D-Studio} \cite{zhao2020hand}, \textbf{SURREAL} \cite{varol2017learning} $\rightarrow$ \textbf{Leeds Sports Pose} \cite{johnson2010clustered}, and \textbf{Synthetic Animal Dataset} \cite{mu2020learning} $\rightarrow$ \textbf{TigDog Dataset} \cite{del2015articulated}. %In addition, we evaluate a challenging pair of image classification datasets for UDA:  \textbf{MNIST} \cite{deng2012mnist} $\rightarrow$ \textbf{SVHN} \cite{netzer2011reading}. Based on multiple works \cite{french2017self, shu2018dirt, dai2020contrastively, kumar2018co}, it has been established that SVHN $\rightarrow$ MNIST is a much easier task where SOTA results are close to supervised learning. However, MNIST $\rightarrow$ SVHN is a much more challenging task where the common consensus \cite{saito2017asymmetric, russo2018source, chen2019progressive, shu2018dirt} is that each proposed method needs a specific data augmentation such as intensity flipping or image standardization to achieve SOTA results. In a real-world setting, it can be time consuming, non-trivial, and not guaranteed to find a data augmentation that works for a particular dataset. Therefore, it is our goal to demonstrate that implicit stylization works without specific data augmentation. Our method trains with only minimal amounts of data augmentation. %(Refer to Section~\ref{} for details on data augmentations).

\subsection{Implementation Details}
The implementation is done with PyTorch with three random seeds 22, 42, and 102. 

\textbf{Learning parameters.} Following the work of \cite{kim2022unified} we use the following data augmentations to increase generalizability but also to keep the level of generalizability consistent throughout the experiments. 

\begin{itemize}
\item \textbf{Human Pose Data Augmentation.} 
Random data augmentations used: 60 degree rotations, shear (-30, 30), translations (0.05, 0.05), scaling (0.6, 1.3), and 0.25 contrast.

\item \textbf{Hand Pose Data Augmentation.} 
Random data augmentations used: 180 degree rotations, shear (-30, 30), translations (0.05, 0.05), scaling (0.6, 1.3), and 0.25 contrast.

\item \textbf{Animal Pose Data Augmentation.} 
Random data augmentations used: 60 degree rotations, shear (-30, 30), translations (0.05, 0.05), scaling (0.6, 1.3), and 0.25 contrast.

\item \textbf{Digits Data Augmentation.} 
Random data augmentations used: rotations, translations, scaling, adjust brightness, adjust contrast. 
\end{itemize}

\textbf{Learning parameters.} Adam with base learning rate of $1e^{-4}$ and learning rate decay of 0.1 at 45 and 60 epochs is chosen as the optimizer for training. 

\textbf{Pre-training.}
Source-only training is done for 40 epochs.

\textbf{Compute Infrastructure.} A batch size of 32 for pose estimation is fed to a single NVIDIA A100 GPU for accelerated training with AMD Milan 7413 CPU available via the shared high performance computing infrastructure. 

\textbf{Hyperparameters.} 
Standard ResNet-101 is used as the backbone and three blocks of upsampling blocks with 256 channels (comprised of 2D transpose convolution, 2D batch norm, ReLU) are used in the decoder. The whole network is trained for 70 epochs.

\subsection{Evaluation Metrics}
The standard metric for 2D pose estimation is Percentage of Correct Keypoint (PCK). Following the work of \citet{kim2022unified} and \citet{li2021synthetic}, all experiments are reported with PCK@0.05 that measures the ratio of correct keypoints that are within 5\% of the image resolution. 

%For \textbf{image classification}, we report on the overall accuracy without class balancing following recent recommendations \cite{french2017self, shu2018dirt, dai2020contrastively, kumar2018co}.
\subsection{Main Results}
In Table~\ref{tab:mainresults}, we compare our implicit stylization model (ImSty) with SOTA models for UDA pose estimation on three sets of 2D pose estimation datasets for domain adaptation. On average over the three datasets, we achieve 0.4\% improvements in PCK@0.05. More importantly , it is further demonstrated in Table~\ref{tab:computation} that implicit stylization reduces the required computation by over 99.99\% and completely removes the number of trainable parameters needed for domain merging compared against the current SOTA UDA pose estimation \cite{kim2022unified}. Qualitative evaluation on the superiority of ImSty can be found in Figure~\ref{fig:qualitativeeval}. %In addition, Table~\ref{tab:digits} demonstrates the superior performance of our implicit stylization method in image classification compared to both generative and non-generative methods given minimal data augmentations. 

As shown in Figure~\ref{fig:tsnediagram}, we employ TSNE \cite{van2008visualizing} to visualize the similarity between feature representations from the source and target domains of the human pose and hand pose datasets. The plots do not provide sufficient evidence of either stylization approach, demonstrating the ability to produce a more  homogeneous distribution (with respect to the invariability of source and target domains) of feature map representations. Therefore, these visualizations further reveal the ineffectiveness of the input-level alignment via stylization used in \citet{kim2022unified}. %While there are differences in homogeneity across datasets, feature maps from RenderedHandPose show slightly more invariance than those of SURREAL in both the explicit and implicit cases.

% \begin{figure}[t]
%     \centering
%             \includegraphics[width=0.5\textwidth]{tsne_uda_icml_1.png}
%             \caption{\textbf{TSNE comparison between \cite{kim2022unified} and our proposed ImSty method.} Visual inspection of the TSNE clustering between the input stylization in \cite{kim2022unified} has close to no difference with our proposed ImSty method. This suggests that ImSty can achieve similar domain alignment compared to the input stylization used in \cite{kim2022unified}.}
%     \label{fig:tsnediagram}
% \end{figure}

\subsection{Ablation Studies}
The main goal of this work is to investigate the role of generative modeling/stylization for domain adaptation in regression. In Table~\ref{table:3}, we compare \citet{kim2022unified} with no stylization and our proposed method which adds Stylization applied to pose estimation offers lackluster results as we see a maximum difference of only two percent between stylized and unstylized accuracy. This suggests a need to revisit the role of domain alignment in the mean-teacher training scheme. %From Table~\ref{table:3}, we see a clear distinction between the importance of stylization for classification tasks versus pose estimation.

%However, when applied to a classification task, the impact of stylization cannot be understated. When applying ImSty to the MNIST $\to$ SVHN task, we see a considerable improvement of 19.8\% against the SOTA generative method increasing from 38.0\% to 57.8\%. This improvement demonstrates the ability of ImSty to achieve theses results without any specific data augmentation. 

\section{Discussion and Conclusion}

\begin{table}[t]
    \centering
    %  \vspace{-20pt}
    \caption{\textbf{Role of stylization in pose estimation} Observe the minimal difference between pose estimation accuracy with and without stylization. This small difference is apparent across all pose estimation tasks questioning the role of domain alignment in pose estimation. }%However, for MNIST $\to$ SVHN ImSty far outperforms our baseline without stylization highlighting the importance of domain alignment.\\}
    \begin{minipage}{.49\textwidth}
    \resizebox{\textwidth}{!}{%
        \begin{tabular}{@{}cccc@{}}
            \multicolumn{4}{c}{\textbf{Rendered Hand Pose Dataset $\to$Hand-3D-Studio}} \\ \midrule
            Joints & UniDA w.o. Stylization & UniDA & Ours (ImSty) \\ \midrule
            MCP & 86.5 $\pm$ 0.5 & 86.5 $\pm$ 0.6& \textbf{87.1 $\pm$ 0.6} \\
            PIP & 85.2 $\pm$ 0.6 & 84.8 $\pm$ 0.4 & \textbf{85.4 $\pm$ 0.2} \\
            DIP & 78.1 $\pm$ 0.4 & 78.7 $\pm$ 0.7 & \textbf{78.8 $\pm$ 0.2} \\
            Fingertip & 66.6 $\pm$ 0.3 & 67.9 $\pm$ 0.6 & \textbf{68.4 $\pm$ 0.3} \\
            All & 79.1 $\pm$ 0.4 & 79.3 $\pm$ 0.5 & \textbf{80.0 $\pm$ 0.1} \\ 
            \bottomrule
        \end{tabular}%
        
        }
        \vspace{16pt}
    \end{minipage}
    \begin{minipage}{.49\textwidth}
    \resizebox{\textwidth}{!}{%
        \begin{tabular}{@{}cccc@{}}
             &  &  \\
            \multicolumn{4}{c}{\textbf{SURREAL $\to$ Leeds Sports Pose}} \\ \midrule
            Joints & UniDA w.o. Stylization & UniDA & Ours (ImSty) \\ \midrule
            Shd & 66.3 $\pm$ 1.3 & \textbf{68.1 $\pm$ 0.3} & 67.9 $\pm$ 3.1 \\
            Elb & 83.1 $\pm$ 1.8 & 84.0 $\pm$ 1.4 & \textbf{84.5 $\pm$ 1.2} \\
            Wrist & 80.7 $\pm$ 1.8 & 82.1 $\pm$ 1.0& \textbf{82.4 $\pm$ 1.3} \\
            Hip & 84.8 $\pm$ 0.3 & \textbf{85.0 $\pm$ 1.3}& 84.9 $\pm$ 0.2 \\
            Knee & 83.4 $\pm$ 0.8 & 83.8 $\pm$ 1.1& \textbf{85.0 $\pm$ 1.1} \\
            Ankle & 83.1 $\pm$ 1.3 & 83.5 $\pm$ 0.7& \textbf{84.4 $\pm$ 0.9} \\
            All & 80.2 $\pm$ 0.7 & 81.1 $\pm$ 0.8 & \textbf{81.5 $\pm$ 1.2} \\ \bottomrule
             &  &  \\
            % \multicolumn{3}{c}{\textbf{MNIST $\to$ SVHN}} \\ \midrule
            % Method & Ours w.o. stylization & Ours (ImSty) \\ \midrule
            % All & 38.0 $\pm$ 4.0 & \textbf{57.8 $\pm$ 3.3} \\ \bottomrule
        \end{tabular}%
        }
    \end{minipage}
    \resizebox{\textwidth}{!}{%
        \begin{tabular}{@{}ccccccc@{}}
             &  &  &  & & & \\
            \multicolumn{7}{c}{\textbf{Synthetic Animal Dataset $\to$ TigDog Dataset}} \\ \midrule
            \multirow{2}{*}{Joints} & \multicolumn{3}{c|}{Horse} & \multicolumn{3}{c}{Tiger} \\ \cmidrule(l){2-7} 
             & UniDA w.o. Stylization & UniDA & \multicolumn{1}{c|}{Ours (ImSty)} & UniDA w.o. Stylization & UniDA & Ours (ImSty) \\ \midrule
            Eye & 89.8 $\pm$ 0.5 &\textbf{91.0 $\pm$ 0.5} & \multicolumn{1}{c|}{90.9 $\pm$ 1.2} & 97.2 $\pm$ 1.0 & 97.3 $\pm$ 0.9&\textbf{97.3 $\pm$ 0.5} \\
            Chin & 92.8 $\pm$ 0.2 &\textbf{93.1 $\pm$ 0.5}& \multicolumn{1}{c|}{92.7 $\pm$ 0.2} & 95.6 $\pm$ 0.2 & \textbf{96.4 $\pm$ 0.6}&96.1 $\pm$ 0.2 \\
            Hoof & \textbf{67.5 $\pm$ 0.7} &65.7 $\pm$ 1.9& \multicolumn{1}{c|}{65.5 $\pm$ 0.6} & 70.7 $\pm$ 0.8 & 70.2 $\pm$ 1.9&\textbf{71.2 $\pm$ 1.2} \\
            Hip & 77.6 $\pm$ 0.9& \textbf{81.2 $\pm$ 1.1}& \multicolumn{1}{c|}{75.6 $\pm$ 1.5} & \textbf{72.7 $\pm$ 2.6} & 69.5 $\pm$ 2.6& 71.5 $\pm$ 1.1 \\
            Knee & \textbf{76.3 $\pm$ 0.8} & 76.1 $\pm$ 1.0& \multicolumn{1}{c|}{75.7 $\pm$ 0.8} & \textbf{61.1 $\pm$ 0.3} & 60.5 $\pm$ 1.9 & 60.6 $\pm$ 0.8 \\
            Shd & \textbf{72.2 $\pm$ 1.1} & 71.6 $\pm$ 1.6& \multicolumn{1}{c|}{71.6 $\pm$ 1.2} & 49.2 $\pm$ 0.8 & 48.8 $\pm$ 4.0 &\textbf{51.8 $\pm$ 0.9} \\
            Elb & 72.4 $\pm$ 1.0 & 72.7 $\pm$ 1.8& \multicolumn{1}{c|}{\textbf{73.9 $\pm$ 0.7}} & 51.0 $\pm$ 1.7 & \textbf{52.3 $\pm$ 0.8} &50.8 $\pm$ 1.5 \\
            All & \textbf{75.5 $\pm$ 0.4} & 75.4 $\pm$ 0.2& \multicolumn{1}{c|}{75.2 $\pm$ 0.2} & 66.3 $\pm$ 0.8 & 66.2 $\pm$ 1.4 &\textbf{66.5 $\pm$ 0.8} \\ \bottomrule
        \end{tabular}%
    }
    \label{table:3}
    \vspace{-10pt}
\end{table}

Domain alignment is an essential part of UDA, shifting the distribution of source and target domains closer to each other. Given the necessity of domain alignment in classification tasks, and the popularity that domain alignment is gaining over the years, it is crucial to investigate the importance of input-level alignment in the mean-teacher scheme for pose estimation (regression) since it requires significantly more computation than feature-level alignment. Moreover, as supported by \cite{cortes2011domain}, many of the observations and assumptions made in classification task do not translate to regression. Hence, there is a clear need to explicitly compare the effect of input-level alignment and feature-level alignment in the regression domain. 

To alleviate that problem, we take a SOTA work on UDA pose estimation \cite{kim2022unified} that utilizes input-level alignment \cite{huang2017arbitrary}, and replace that with the AdaIN module which does not require any learnable parameters and minimal computations (demonstrated in Figure~\ref{fig:scatter} and Table~\ref{tab:computation}). By achieving SOTA results in UDA pose estimation, we demonstrate that input-level alignment via generative methods and stylization may not be necessary in regression tasks. The resulting impact \textbf{reduces the work, time, and computational cost required for the generative model by a significant margin}. In addition, any instability caused by generative adversarial networks can be avoided with implicit stylization. Finally, this opens doors for use of domain adaptation in data scarce regimes.

\subsection{Limitations and Future Work}
Although ImSty achieved SOTA performance on regression tasks, while significantly reducing the amount of computation and trainable parameters (by replacing the input-level alignment model with a new feature-level alignment module), the effect on regression remains small. This raises an interesting question about the role and limitations of domain alignment in the mean-teacher training scheme for various machine learning tasks, which we aim to investigate next.

\subsection{Broader Impacts} Computer vision research which focuses on human pose estimation can be used for surveillance, which raises privacy invasion and human rights concerns. Researchers and practitioners both need to be educated and informed about reliance on these technologies regarding these potential risks. Privacy preserving machine learning offers a way to mitigate these risks, and must be considered for real-world deployment. On the other hand, there is also a need to develop legal protections and regulations for users and various entities.

{\small
\bibliographystyle{unsrtnat}
\bibliography{neurips_2023}
}

%%%%%%%%%%%%%%%%%%%%%%%%%%%%%%%%%%%%%%%%%%%%%%%%%%%%%%%%%%%%

\end{document}

\maketitle

\appendix

\section{Digits Classification}
\label{app:otherexp}
\subsection{Methodology}

Implicit stylization for UDA in image classification shares a similar pipeline to that of pose estimation. There are two main differences: replacement of the decoder with a classification head, and replacement of the batch-wise masking with an adaptive confidence threshold based on sample prediction probabilities. 

Similar to pose estimation, given a labeled classification dataset from the source domain $\sS = \{(x^{i}_{s}, y^{s}_{s})\}$  with images $\sX_{s}$ and corresponding classification labels $y_{s}$, along with unlabeled classification dataset from the target domain $\sT = \{x^{i}_{t} \}$, the goal is to generalize a model $h$ to the target domain $\sT$ based on learning from the source domain $\sS$.

Given the implicit style transformed feature maps 
$\tF_{s \rightarrow t}$ and $\tF_{t \rightarrow s}$ from \eqref{eq:reversenormalization} (we use LeNet-5 \cite{lecun1998gradient} with ReLU activations and batch normalization following \cite{ganin2015unsupervised, russo2018source}), $\tF_{s \rightarrow t}$ and $\tF_{t}$ are passed to the student classification head $H_{\text{stu}}$ that linearly maps the feature maps to $C$ classes: 
\begin{equation}
\begin{aligned}
    \hat{y}_{s} &= H_{\text{stu}}(\tF_{s \rightarrow t}) \in \mathbb{R}^{C}\\
     \hat{y}_{t} &= H_{\text{stu}}(\tF_{t}) \in \mathbb{R}^{C}
    \end{aligned}
    \label{eq:decoder1}
\end{equation}
Then, $\tF_{t \rightarrow s}$ is passed to the teacher classification head $H_{\text{tea}}$ to generate pseudo-labels $y_{t}$:
\begin{equation}
\begin{aligned}
     y_{t} &= H_{\text{tea}}(\tF_{t \rightarrow s})
    \end{aligned}
    \label{eq:decode2r}
\end{equation}
The maximum probability $\text{max}(\hat{y}_{\{s,t\}})$ for the whole training population $N$ is collected to determine the confidence threshold $\beta$ for the subsequent epoch. $\beta$ is determined by finding the $p\text{ th}$ percentile value from the $N$ probabilities. In the following epoch, all the pseudo-labels $y_{\text{pseudo}}$ with $\text{max}(\hat{y}_{\text{pseudo}}) < \beta $ are masked as zeros in the loss function. We initialize $\beta$ with zero and choose $p$ as 50\% for our experiments. Binary cross entropy (BCE) loss is used as the loss function for both supervised $\mathcal{L}_{\text{sup}}$ and unsupervised $\mathcal{L}_{\text{unsup}}$ loss with a weighting factor $\lambda$ for the overall loss $\mathcal{L}_{\text{total}}$:
\begin{equation}
\begin{aligned}
    \mathcal{L}_{\text{sup}} &= \text{BCE}(y_{s}, \hat{y}_{s }) \\
    \mathcal{L}_{\text{unsup}} &= \text{BCE}(y_{t}, \hat{y}_{t}) \\
    \mathcal{L}_{\text{total}} &= \mathcal{L}_{\text{sup}}+\lambda \mathcal{L}_{\text{unsup}}
    \end{aligned}
    \label{eq:loss}
\end{equation}
\subsection{Experiments}
We evaluate a challenging pair of image classification datasets for UDA:  \textbf{MNIST} \cite{deng2012mnist} $\rightarrow$ \textbf{SVHN} \cite{netzer2011reading}. Based on multiple works \cite{french2017self, shu2018dirt, dai2020contrastively, kumar2018co}, it has been established that SVHN $\rightarrow$ MNIST is a much easier task where SOTA results are close to supervised learning. However, MNIST $\rightarrow$ SVHN is a much more challenging task where the common consensus \cite{saito2017asymmetric, russo2018source, chen2019progressive, shu2018dirt} is that each proposed method needs a specific data augmentation such as intensity flipping or image standardization to achieve SOTA results. In a real-world setting, it can be time consuming, non-trivial, and not guaranteed to find a data augmentation that works for a particular dataset. Therefore, it is our goal to demonstrate that implicit stylization works without specific data augmentation. Our method trains with only minimal amounts of data augmentation.

\textbf{Evaluation Metrics.} We report on the overall accuracy without class balancing following recent recommendations \cite{french2017self, shu2018dirt, dai2020contrastively, kumar2018co}.
\begin{table*}[t]
    \centering
    \caption{\textbf{Digit classification results on MNIST $\rightarrow$ SVHN.} It is well established from \cite{french2017self} that MNIST $\rightarrow$ SVHN is much more challenging than SVHN $\rightarrow$ MNIST. In addition, the results of many works vary greatly depending on the specific data augmentation. Since manually searching for data augmentations that work for the target domain is costly and time-consuming, we compare SOTA methods on minimal data augmentations. Observe that given minimal data augmentations, our method achieves SOTA results on UDA digit classification without generative methods. "SBADA-GAN$^{R}$" indicates reproduced values.}
    \begin{tabular}{ccc}
    \toprule 
         Method & Accuracy & Generative\\
         \midrule 
         RevGrad \cite{ganin2015unsupervised} & 35.7 & \\
         DCRN \cite{ghifary2016deep} & 40.1 &\\
         G2A \cite{sankaranarayanan2018generate} & 36.4 & \checkmark\\
         SE \cite{french2017self} & 37.5 &\\
         ATT \cite{saito2017asymmetric} & 52.8 &\\
         SBADA-GAN$^{R}$ \cite{russo2018source} & 47.9 $\pm$ 1.7&\checkmark\\
         % SBADA-GAN\cite{russo2018source} & \textbf{61.1} &\checkmark\\
         PFAN \cite{chen2019progressive} & 57.6 $\pm$ 1.8 &\\
         DIRT-T \cite{shu2018dirt} & 54.5 & \\
         Ours (ImSty) & 	\textbf{57.82} $\pm$ 3.2 & \\
         \bottomrule
    \end{tabular}
    \label{tab:digits}
\end{table*}

\textbf{Results.} Table~\ref{tab:digits} demonstrates the superior performance of our implicit stylization method in image classification compared to both generative and non-generative methods given minimal data augmentations. 

When applying ImSty to the MNIST $\to$ SVHN task, we see a considerable improvement of 19.8\% against the SOTA generative method increasing from 38.0\% to 57.8\%. This improvement demonstrates the ability of ImSty to achieve theses results without any specific data augmentation. 

{\small
\bibliographystyle{unsrtnat}
\bibliography{neurips_2023}
}